\documentclass[11pt]{article}

\usepackage[final]{acl}

\usepackage{times}
\usepackage{latexsym}

\usepackage[T1]{fontenc}

\usepackage[utf8]{inputenc}

\usepackage{microtype}

\usepackage{inconsolata}

\usepackage{graphicx}

\usepackage[utf8]{inputenc} 
\usepackage[T1]{fontenc}    
\usepackage{hyperref}       
\usepackage{url}            
\usepackage{booktabs}       
\usepackage{amsfonts}       
\usepackage{nicefrac}       
\usepackage{microtype}      
\usepackage{xcolor}         

\usepackage{microtype}
\usepackage{hyperref}
\usepackage{url}
\usepackage{booktabs}

\usepackage{lineno}
\usepackage{graphicx}
\usepackage{subcaption}
\usepackage{wrapfig}

\usepackage{amsmath}
\usepackage{amssymb}
\usepackage{mathtools}
\usepackage{amsthm}
\usepackage{multirow}

\theoremstyle{plain}

\definecolor{darkblue}{rgb}{0, 0, 0.5}
\hypersetup{colorlinks=true, citecolor=darkblue, linkcolor=darkblue, urlcolor=darkblue}

\usepackage{multirow}
\usepackage{colortbl}
\usepackage{xspace}

\newcommand{\metric}{CUD\xspace}
\newcommand{\eg}{\textit{e.g.,}\xspace}
\newcommand{\ie}{\textit{i.e.,}\xspace}

\definecolor{Gray}{gray}{0.9}
\definecolor{lightred}{RGB}{170,50,50}
\definecolor{lightgreen}{RGB}{0,140,0}

\newcommand{\xhdr}[1]{\vspace{0em}\noindent{{\bf #1:}}}

\title{A Mechanistic Perspective and Circuit-Guided Difficulty Metric\\for Unlearning}

\author{Jiali Cheng \\
  University of Massachusetts Lowell \\
  \texttt{jiali\_cheng@uml.edu} \\\And
  Ziheng Chen \\
  Walmart Global Tech\\
  \texttt{albertchen1993pokemon@gmail.com} \\\AND
  Chirag Agarwal \\
  University of Virginia \\
  \texttt{chiragagarwal@virginia.edu} \\\And
  Hadi Amiri \\
  University of Massachusetts Lowell \\
  \texttt{hadi\_amiri@uml.edu} \\
}

\begin{document}
\maketitle


\begin{abstract}
Machine unlearning is becoming essential for building trustworthy and compliant language models. 
Yet unlearning success varies considerably across individual samples: some are reliably erased, while others persist despite the same procedure. 
%
We argue that this disparity is not only a data-side phenomenon, but also reflects model-internal mechanisms that encode and protect memorized information. 
We study this problem from a mechanistic perspective based on model circuits--structured interaction pathways that govern how predictions are formed. 
We propose Circuit-guided Unlearning Difficulty (\metric), a {\em pre-unlearning} metric that assigns each sample a continuous difficulty score using circuit-level signals. 
%
Extensive experiments demonstrate that \metric reliably separates intrinsically easy and hard samples, and remains stable across unlearning methods.
We identify key circuit-level patterns that reveal a mechanistic signature of unlearning difficulty: easy-to-unlearn samples are associated with shorter, shallower interactions concentrated in earlier-to-intermediate parts of the original model, whereas hard-to-unlearn samples rely on longer and deeper pathways closer to late-stage computation. 
%
Compared to existing qualitative studies, \metric takes a first step toward a principled, fine-grained, and interpretable analysis of unlearning difficulty; and motivates the development of unlearning methods grounded in model mechanisms.\looseness-1
\end{abstract}
\section{Introduction}
Machine unlearning is the process of removing the knowledge of specific training data (e.g., noisy or proprietary data) from a trained model without retraining from scratch~\citep{cao2015towards,bourtoule2021machine,liu2024rethinking,jia-etal-2024-soul}. This need is driven by both legal and ethical imperatives, such as removing copyrighted data from large language models (LLMs)~\citep{eldan2023whos,shi2025muse}, as well as practical necessity of purging outdated or incorrect information~\citep{dhingra-etal-2022-time,cheng2025tool}. As LLMs scale in size and training cost, understanding and interpreting unlearning methods is becoming an important research frontier.

There is growing evidence that unlearning performance varies substantially across individual samples: some examples are erased reliably, while others persist despite unlearning~\citep{fan2024challenging,ebrahimpour-boroojeny2025not,hong-etal-2025-intrinsic,wei2025llms,rizwan2024instance}. Although recent unlearning methods have made notable progress, why this disparity arises is poorly understood. Existing work largely treats unlearning difficulty as a data-side phenomenon, attributing failures to spurious correlations, redundancy, or dataset structure~\citep{zhao2024what,krishnan2025not}. As a result, we still lack a mechanistic, model-internal explanation of \textit{why certain samples are intrinsically harder to forget}.

In this work, we aim to mitigate the above gap by addressing two research questions: 
\textbf{RQ1:} {\em can we define and quantify a sample's unlearning difficulty from a mechanistic perspective before unlearning?} 
\textbf{RQ2:} {\em can this predicted difficulty be linked to the internal mechanisms (e.g. circuits) within the model?}
We perform circuit-level analysis~\citep{conmy2023towards,hanna2024have,haklay-etal-2025-position} of LLM unlearning. 
We show that easy- and hard-to-unlearn samples are memorized through structurally different internal circuits, leading to distinct post-unlearning behaviors. 
Building on these insights, we introduce a circuit-guided metric that quantifies unlearning difficulty directly from model internals, independent of the unlearning algorithm itself. 
Our contributions are:

\begin{itemize}
\itemsep0pt
    \item the first circuit-level analysis of disparity in unlearning performance, which reveals how different internal structures underpin unlearning of easy and hard samples, and
    \item circuit-guided unlearning difficulty (\metric) score--a mechanistic metric that measures the intrinsic unlearning difficulty of samples.
\end{itemize}

We demonstrate that easy- and hard-to-unlearn samples are intrinsically distinct at the circuit level: easy samples primarily activate shallower edges, while hard forget samples depend on deeper edges. Leveraging \metric, we construct intrinsically easy and hard forget sets, whose difficulty is demonstrated through unlearning effectiveness of the same unlearning algorithm. Across five unlearning methods, hard forget sets lead to a drop of 14.1 points in unlearning effectiveness, while easy forget sets yield an improvement of 3.3 points.
\section{Background and Related Work}
Our work is at the intersection of explainability and unlearning. Below, we discuss the related works.\looseness-1

\paragraph{Explainability in Unlearning Performances:}
\citet{fan2024salun}, \citet{jia2024wagle} find that there is a subset of parameters that are prominent to forget set, identified by gradient-based Saliency Map~\citep{simonyan2013deep}. Only optimizing this subset of parameters leads to significant performance advantage in unlearning. 
\cite{chen2025clue,chen2026cure} find that forget and retain samples correlate with different circuits in the model.
\citet{hong2024intrinsic} find that unlearning performance is correlated with the parametric concept vectors discovered in the MLP layers of the model. 
Recent work discovers loss re-weightting as an effective approach of unlearning, some tokens are more forget-related, and some are less relevant~\citep{yang2025exploring,wan2025not}.
Other work finds that smooth loss landscape can lead to more robust unlearning against adversarial attacks~\citep{cheng2025understanding,fan2025towards}.
\textit{However, existing work generally lacks mechanistic explanations in unlearning, especially the distinct unlearning performances across different samples.}\looseness-1



\paragraph{Explainability in Unlearning Difficulty:} Earlier works use heuristics to find hard-to-unlearn samples, for example, samples that are close to test set are hard to unlearn~\citep{cheng2023gnndelete,chen2025frog,wei2025llms}.
Additionally, samples that are 1) highly memorized, and 2) deeply entangled with the retain set in embedding space are generally hard to unlearn~\citep{zhao2024what}.
Later, \citet{fan2024challenging} propose a principled optimization strategy to find hard-to-unlearn samples, \ie samples with low loss (high memorization) post-unlearning. 
\citet{cheng2025understanding} argue that if an unlearning task has a smooth loss landscape, the task is easy. 
Recent work discovers coreset effect -- unlearning the core forget set is equivalent to unlearning the entire forget set~\citep{fan2024challenging,patil2025upcore,pal2025llm}. This is due to shared key tokens in the forget set, outliers, and similarity to the retain samples.

\paragraph{Circuit Finding:}\label{sec:circuit}

Let $a$ denote some activation in the computational graph and an edge $e{=}(a_u \rightarrow a_v)$ connect an upstream activation $a_u$ to a downstream activation $a_v$. Given a metric $M$, \eg output probability of the model, the importance of $a$ is measured using the change of $M$ when using clean input $x_{\text{clean}}$ and patch input $x_{\text{patch}}$~\citep{NEURIPS2020_92650b2e,finlayson-etal-2021-causal,meng2022locating,marks2024sparse}, \ie
\begin{equation}
M\!\left(x_{\text{clean}} \mid \mathrm{do}(a = a_{\text{patch}})\right) - M(x_{\text{clean}}),
\end{equation} where $a_{\text{patch}}$ denotes the activation of $a$ when using $x_{\text{patch}}$ as input.
For example, $x_{\text{clean}}{=}$ The cat ``is'' sitting on the mat, and $x_{\text{patch}}{=}$ The cat ``are'' sitting on the mat. We measure the change of output probability as a proxy of how important $a$ is to explaining the behavior or using singular or plural verbs. When $x_{\text{patch}}$ is not available or not easy to obtain, we can use zero-ablation, \ie $a_{\text{patch}} = 0$~\citep{hanna2024have,marks2024sparse}.

Edge Attribution Patching (EAP)~\citep{nanda2023attribution,syed2024attribution} approximates the indirect causal effect of an edge via a first-order linearization of the metric around the clean input:
\begin{equation}\label{eq:eap}
    \text{EAP}(e) = \Delta a_u \cdot \frac{\partial m(x_{\text{clean}})}{\partial a_v}.
\end{equation}
This formulation enables efficient and scalable circuit discovery using a small number of steps, but relies on a single local gradient and may underestimate edges involved in nonlinear or saturated computations. To obtain a more faithful attribution, EAP with Integrated Gradients (EAP-IG)~\citep{hanna2024have} replaces the single-point gradient with an integrated gradient along a linear interpolation path between clean and patched inputs:
\begin{equation}
    x(\alpha) = x_{\text{clean}} + \alpha \big(x_{\text{patch}} - x_{\text{clean}}\big), \quad \alpha \in [0,1].
\end{equation}
The EAP-IG score for edge $e$ is then defined as:
\begin{equation}\label{eq:eapig}
\text{EAP-IG}(e) = \Delta a_u \cdot \int_{0}^{1} \frac{\partial m(x(\alpha))}{\partial a_v} d \alpha.
\end{equation}
\begin{figure*}
    \centering
    \vspace{-20pt}
    \includegraphics[width=\linewidth]{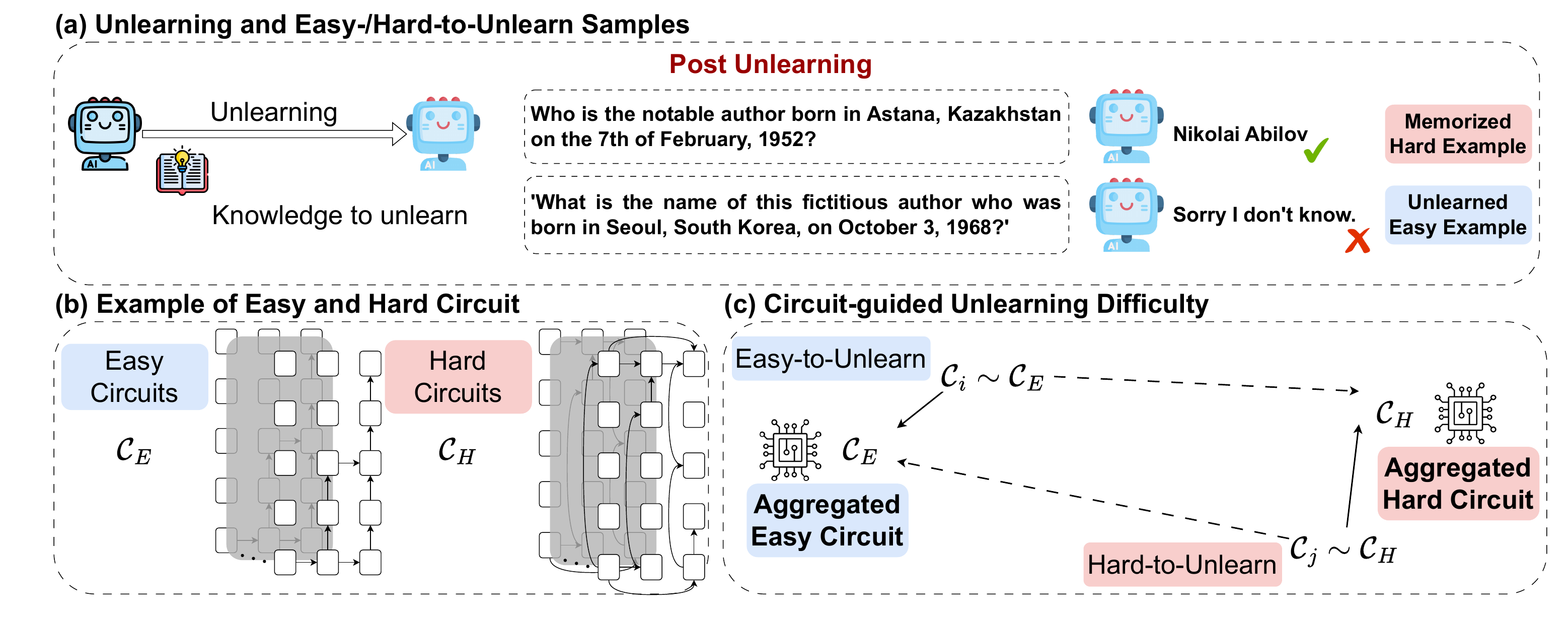}
    \caption{A mechanistic perspective of unlearning difficulty. \textbf{(a)} Illustration of unlearning and post-unlearning performance differences. \textbf{(b)} An example of mechanistic differences of circuits between easy- and hard-to-unlearn samples. \textbf{(c)} Proposed Circuit-guided Unlearning Difficulty (\metric) score to quantify unlearning difficulty of samples.\looseness-1}
    \label{fig:fig1}
\end{figure*}

\section{A Mechanistic Perspective on Unlearning Difficulty}
We introduce Circuit-guided Unlearning Difficulty (\metric) score -- a principled metric to quantify the mechanistic unlearning difficulty of individual samples. \metric enables systematic evaluation of unlearning difficulty at the sample level and facilitates the construction of challenging forget sets for stress-testing unlearning algorithms. Our approach consists of 1) finding two reference circuits (easy and hard) and 2) measuring the similarity of the query sample to the reference circuits.

\paragraph{Notation:} Let $f_o$ be a model parametrized by $\theta$ trained on dataset $\mathcal{D}$ with task loss $L$. In addition, assume that $\mathcal{D}$ can be divided into two disjoint sets: the forget set $\mathcal{D}_f$ and the retain set $\mathcal{D}_r = \mathcal{D} \setminus \mathcal{D}_f$.


\subsection{Finding Reference Circuits}
Each sample $z_i \in \mathcal{D}_f$ encodes a distinct unit of knowledge to be forgotten. For each $z_i$, we first extract its corresponding circuit from the original model $f_o$, denoted as $\mathcal{C}_i$, using the method in \S\ref{sec:circuit}. The circuit is represented as a structured matrix capturing the functional interactions among model components that are responsible for the prediction on $z_i$. We anchor the unlearning difficulty of $z_i$ against two reference circuits: an \emph{easy-to-unlearn} anchor $\mathcal{C}_E$ and a \emph{hard-to-unlearn} anchor $\mathcal{C}_H$. Intuitively, difficulty of unlearning is determined by where $\mathcal{C}_i$ lies along the spectrum spanned by these two anchors.

Following \citet{fan2024challenging}, we use bi-level optimization objective to find easy and hard to unlearn samples. We define a binary mask $\mathbf{w} = \{0, 1\}, |\mathbf{w}| = |\mathcal{D}_f|$ to indicate which samples are in the forget set, \ie $w_i = 1$ represents $z_i \in \mathcal{D}_f$.
We employ $\mathcal{w}$ to select which samples to include in the corresponding set:\looseness-1
\begin{multline}\label{eq:find_easy}
    \max_{\mathbf{w}} 
    \sum_{z_i \in \mathcal{D}_f} \big[ w_i \, L \big(z_i; \theta_u(\mathbf{w})\big) \big]
    \;+\; \lambda \lVert \mathbf{w} \rVert_2^2, \quad s.t. \\
    \theta_u(\mathbf{w}) = \arg\min_{\theta} \; L_{\mathrm{MU}}(\theta; \mathbf{w}),
\end{multline}
\begin{multline}\label{eq:find_hard}
    \min_{\mathbf{w}} 
    \sum_{z_i \in \mathcal{D}_f} \big[ w_i \, L \big(z_i; \theta_u(\mathbf{w})\big) \big]
    \;+\; \lambda \lVert \mathbf{w} \rVert_2^2, \quad s.t. \\
    \theta_u(\mathbf{w}) = \arg\min_{\theta} \; L_{\mathrm{MU}}(\theta; \mathbf{w}),
\end{multline} where $L_{\text{MU}} = \sum_{z_i \in \mathcal{D}_r} L(z_i) - \sum_{z_j \in \mathcal{D}_f} w_jL(z_j)$ -- 
a straightforward unlearning formulation, $\theta_u$ denotes the model parameters post-unlearning, and $\lambda$ is a hyperparameter that encourages selecting a small set of samples. 

Intuitively, Eq.~\ref{eq:find_easy} finds samples that have increased loss (\ie low memorization, easy-to-unlearn) post-unlearning and are thus considered easier to forget, denoted as $\mathcal{D}_{f,E}$. While Eq.~\ref{eq:find_hard} finds samples that remain low loss (\ie high memorization, hard-to-unlearn) post-unlearning, where unlearning remains unsuccessful, denoted as $\mathcal{D}_{f,H}$.\looseness-1

To account for any bias or stochasticity, we repeat each unlearning method five times with different seeds and take the common samples in all runs to get the stable $\mathcal{D}_{f,E}, \mathcal{D}_{f,H}$.

After that, we use circuit finding methods to locate the circuits for $\mathcal{D}_{f,E}, \mathcal{D}_{f,H}$ on the original model $f_o$, denoted as $\mathcal{C}_E, \mathcal{C}_H$, respectively.

\subsection{\metric Score}
We represent each circuit as a binary matrix of edges, where element $\mathcal{C}[i, j] = 1$ if the corresponding edge appears in the circuit. We flatten each circuit matrix into a vector representation
and then compute the similarities of the sample circuit to the easy anchor $\mathcal{C}_E$ and hard anchor $\mathcal{C}_H$, \ie
\begin{equation}
\begin{aligned}
s_E &= \operatorname{sim} \big(\operatorname{vec}(\mathcal{C}_i), \operatorname{vec}(\mathcal{C}_E)\big),\\
s_H &= \operatorname{sim} \big(\operatorname{vec}(\mathcal{C}_i), \operatorname{vec}(\mathcal{C}_H)\big),
\end{aligned}
\end{equation}
where $\operatorname{sim}(\cdot)$ denotes some similarity metric, and $\operatorname{vec}(\cdot)$ flattens a circuit matrix into a vector. We define the \metric score of sample $z_i$ as:
\begin{equation}
\mathrm{\metric}(z_i) = \frac{1 - s_E} {(1 - s_E) + (1 - s_H)},
\end{equation} which yields a normalized score in $[0,1]$. When $\mathcal{C}_i$ is more similar to $\mathcal{C}_E$ (\ie, $\mathcal{C}_i \rightarrow \mathcal{C}_E$), the unlearning difficulty of $z_i$ will be close to 0, indicating that $z_i$ is easy to unlearn. When $\mathcal{C}_i \rightarrow \mathcal{C}_H$, the unlearning difficulty of $z_i$ will be close to 1, indicating that $z_i$ is hard to unlearn. A larger value of \metric over $\mathcal{D}_f$ samples suggests that the model still retains strong internal signals of $\mathcal{D}_f$ post unlearning, indicating that these samples are harder to erase.

Conceptually, \metric captures the difficulty of unlearning as a relative geometric position in circuit space rather than an outcome-dependent post-hoc measure. As a result, it enables offline estimation of unlearning difficulty prior to applying any unlearning algorithm and supports principled construction of adversarial or curriculum-based unlearning batches.

\begin{figure}
    \centering
    \includegraphics[width=0.95\linewidth]{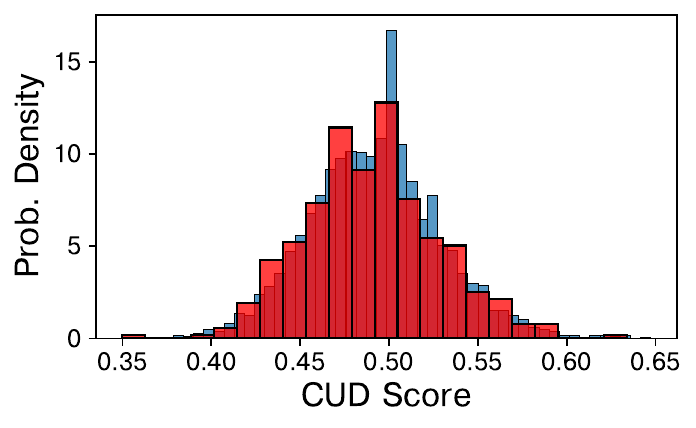}
    \caption{\metric score distribution of samples from TOFU. \textcolor{red}{Red: Default forget set}. \textcolor{blue}{Blue: All samples}. The default forget samples closely matches the overall distribution, with a wide coverage of all difficulty levels, suggesting that TOFU's default forget set represents a mid-level unlearning difficulty. Using \metric, we can select a harder / easier forget sets than the default to control task difficulty and better stress-test unlearning methods.\looseness-1}
    \label{fig:cud_tofu_per_sample}
\end{figure}

\section{Experiments}

\paragraph{Datasets:} We consider the following LLM unlearning benchmarks. On TOFU~\citep{maini2024tofu}, a dataset of fabricated author profiles,  the LLM is required to unlearn personal information about specific authors. We use the \texttt{forget10} split with 400 forget samples. On MocieLens-1M~\citep{wang2025towards}, the LLM is required to forget recommendation information of user-item relationship. We use the forget-retain split in \citet{wang2025towards} with 500 forget samples.



\paragraph{Unlearning Methods:} We consider the following unlearning methods: 1) GradAscent, 2) GradDiff~\citep{maini2024tofu}, 3) NPO~\citep{zhang2024negative}, 4) SimNPO~\citep{fan2024simplicity}, 5) UNDIAL~\citep{dong-etal-2025-undial}, 6) E2UREC~\citep{wang2025towards}, and 7) RecEraser~\citep{chen2022recommendation}. These models are described in Appendix~\ref{sec:ori_model}.

\paragraph{Evaluation Metrics:} Following standard evaluation methods, we report i) ROUGE on TOFU~\citep{maini2024tofu}, ii) AUC
on test set for LLM recommendation, and iii) Jensen--Shannon divergence (JSD) between the predictions of the unlearned model and those of a retrained-from-scratch model, thereby reflecting the effectiveness of the unlearning process. To make results easy to interpret, we follow prior work~\citep{fan2025towards,reisizadeh2025blur,liu2024rethinking} and convert unlearning metric ($\downarrow$) into $\text{unlearning efficacy} (\uparrow) = 1 - \text{unlearning performance}$, where higher is better.

\begin{table*}[t]
\small
\centering
\caption{\metric identifies easy-/hard-to-unlearn forget sets. Under the same unlearning settings, hard set has lower Unlearning efficacy, retain performance, and general knowledge, indicating greater resistance to forgetting, whereas the easy set achieves higher performance across all metrics. Default set: the default forget/retain split on TOFU. Hard set: Hard forget set selected by \metric. Similar for Easy set. Numbers in parenthesis report the gap to default set and $p$-value of difference, respectively. See Tables~\ref{tab:rec_new_split_gpt2}-\ref{tab:rec_new_split_llama} in Appendix~\ref{sec:additional_results} for results on LLMRec unlearning. In the results below, $^*$ denotes a p-value \(\le 0.05\), $^{**}$ denotes a p-value \(\le 0.01\), and $^{***}$ denotes a p-value \(p\le 0.001\). 
}
\label{tab:tofu_new_split}
\begin{tabular}{l|l|ccc}
\toprule
\textbf{Unlearn Method} & \textbf{Choice of $\mathcal{D}_f$} & \textbf{Unlearn Efficacy ($\uparrow$)} & \textbf{Retain ($\uparrow$)} & \textbf{General Knowledge ($\uparrow$)} \\
\midrule
Prior-Unlearn & - & 22.0 & 79.3 & 81.2 \\
\midrule
\multirow{3}{*}{GradDiff} 
& Default Set & 43.4 & 78.3 & 77.4 \\
\cmidrule{2-5}
& Hard Set by \metric & 33.2 (\textcolor{lightred}{-10.2})$^{***}$ & 73.5 (\textcolor{lightred}{-4.8})$^{***}$ & 76.2 (\textcolor{lightred}{-1.2})$^{**}$ \\
& Easy Set by \metric & 50.4 (\textcolor{lightgreen}{+7.0})$^{***}$ & 78.8 (\textcolor{lightgreen}{+0.5})$^{**}$ & 77.6 (\textcolor{lightgreen}{+0.2})$^{*}$ \\
\midrule
\multirow{3}{*}{NPO} 
& Default Set & 54.0 & 76.2 & 79.9 \\
\cmidrule{2-5}
& Hard Set by \metric & 35.7 (\textcolor{lightred}{-18.3})$^{***}$ & 72.8 (\textcolor{lightred}{-3.4})$^{***}$ & 75.5 (\textcolor{lightred}{-4.4})$^{***}$ \\
& Easy Set by \metric & 57.9 (\textcolor{lightgreen}{+3.9})$^{***}$ & 77.5 (\textcolor{lightgreen}{+1.3})$^{**}$ & 77.6 (\textcolor{lightred}{-2.3})$^{**}$ \\
\midrule
\multirow{3}{*}{SimNPO} 
& Default Set & 65.5 & 56.0 & 80.3 \\
\cmidrule{2-5}
& Hard Set by \metric & 47.9 (\textcolor{lightred}{-17.6})$^{***}$ & 55.1 (\textcolor{lightred}{-0.9})$^{**}$ & 77.5 (\textcolor{lightred}{-2.8})$^{***}$ \\
& Easy Set by \metric & 67.3 (\textcolor{lightgreen}{+1.8})$^{**}$ & 56.8 (\textcolor{lightgreen}{+0.8})$^{**}$ & 80.4 (\textcolor{lightgreen}{+0.1})$^{*}$ \\
\midrule
\multirow{3}{*}{UNDIAL} 
& Default Set & 68.3 & 56.3 & 64.2 \\
\cmidrule{2-5}
& Hard Set by \metric & 57.9 (\textcolor{lightred}{-10.4})$^{***}$ & 55.6 (\textcolor{lightred}{-0.7})$^{**}$ & 63.3 (\textcolor{lightred}{-0.9})$^{**}$ \\
& Easy Set by \metric & 68.6 (\textcolor{lightgreen}{+0.3})$^{*}$ & 58.7 (\textcolor{lightgreen}{+2.4})$^{***}$ & 65.5 (\textcolor{lightgreen}{+1.3})$^{**}$ \\
\midrule
\multirow{3}{*}{\textbf{Average}} 
& Default Set & 57.8 & 66.7 & 75.5 \\
\cmidrule{2-5}
& Hard Set by \metric & 43.7 (\textcolor{lightred}{-14.1})$^{***}$ & 64.3 (\textcolor{lightred}{-2.4})$^{***}$ & 73.1 (\textcolor{lightred}{-2.4})$^{***}$ \\
& Easy Set by \metric & 61.1 (\textcolor{lightgreen}{+3.3})$^{***}$ & 68.0 (\textcolor{lightgreen}{+1.3})$^{***}$ & 75.3 (\textcolor{lightred}{-0.2})$^{*}$ \\
\bottomrule
\end{tabular}
\end{table*}

\subsection{Results}
We present the results on \metric score and study two questions: 
1) Does \metric score truly capture the intrinsic difficulty of sample unlearning? 
2) How sensitive is \metric to key design choices?


\paragraph{\metric captures method-independent unlearning difficulty:} Using \metric, we construct \emph{new forget sets} that are intrinsically easier or harder to unlearn than the default forget set. Under the same unlearning setting (method, hyperparameters, etc.), merely replacing the default set with the \metric-selected sets can lead to statistically different unlearning performance, demonstrated in Table~\ref{tab:tofu_new_split}.

Across all unlearning methods, replacing the default TOFU forget set with the \emph{easy set} identified by \metric consistently yields better unlearning efficacy (first column), with improvements ranging from $+1.8$ to $+7.0$ points. These gains are statistically significant in most cases, with average improvements of $+3.3$ points and corresponding $p$-values on the order of $10^{-3}$. Importantly, retain performance and general knowledge accuracy are largely preserved or slightly improved, indicating that easier-to-unlearn samples can be removed more effectively without introducing additional degradation to the model.

In the LLM recommendation setting, unlearning on the easy set shows considerably higher unlearning efficiency, with negligible degradation in the utility of the original recommender. In particular, it outperforms unlearning on a randomly selected forget set by $7.2\%$, see Table~\ref{tab:rec_new_split_gpt2}--\ref{tab:rec_new_split_llama} in Appendix~\ref{sec:additional_results}. In contrast, using the \emph{hard set} selected by \metric consistently results in substantially worse performance than the default split, indicating significantly greater resistance to forgetting. Unlearning efficacy drops up to $18\%$ across all methods, with highly significant differences ($p \leq 10^{-13}$ in all cases). Moreover, unlearning the hard set also leads to systematic degradation in retain and general knowledge metrics, suggesting that these samples are more strongly entangled with the model’s internal representations and therefore harder to remove without collateral effects.

\looseness=-1{
The clear and consistent separation between easy, default, and hard splits confirms that \metric reliably stratifies samples by intrinsic unlearning difficulty: samples in the easy set are indeed easier to forget than those in the default split, whereas samples in the hard set are harder to unlearn.

\paragraph{\metric is robust to similarity metrics:} Table~\ref{tab:tofu_new_split_comparison} in Appendix~\ref{sec:additional_results}, shows that \metric is robust to the choice of similarity metric used in its construction. We instantiate \metric with three different similarity measures (Cosine, Jaccard, and Hamming), and evaluate the resulting easy and hard forget sets under identical unlearning settings.
Across all similarity metrics, \metric consistently induces the same qualitative separation on the default TOFU split. The \emph{hard sets} selected by \metric are uniformly more difficult to unlearn than the default set, showing substantially lower unlearning efficacy with large and statistically significant drops (ranging from $-10.6$ to $-14.1$, all with $p \leq 10^{-13}$). These hard sets also lead to consistent degradation in retain performance and general knowledge, indicating stronger resistance to forgetting and greater interference with the model’s internal representations.

\looseness=-1 Conversely, the \emph{easy sets} identified by \metric consistently outperform the default split across all similarity metrics. Unlearning efficacy improves by $+2.4$ to $+3.7$ points with statistically significant gains, while retain performance is preserved or modestly improved. General knowledge accuracy remains largely unchanged, with differences that are small and statistically insignificant, suggesting that easier-to-unlearn samples can be removed more cleanly regardless of the similarity metric used. The close quantitative agreement across Cosine, Jaccard, and Hamming variants demonstrates that \metric’s ability to stratify unlearning difficulty is not sensitive to a particular design choice. Instead, \metric captures a stable, intrinsic notion of unlearning difficulty that persists across different similarity formulations, reinforcing its robustness and practical applicability.

\begin{figure}
    \centering
    \includegraphics[width=0.98\linewidth]{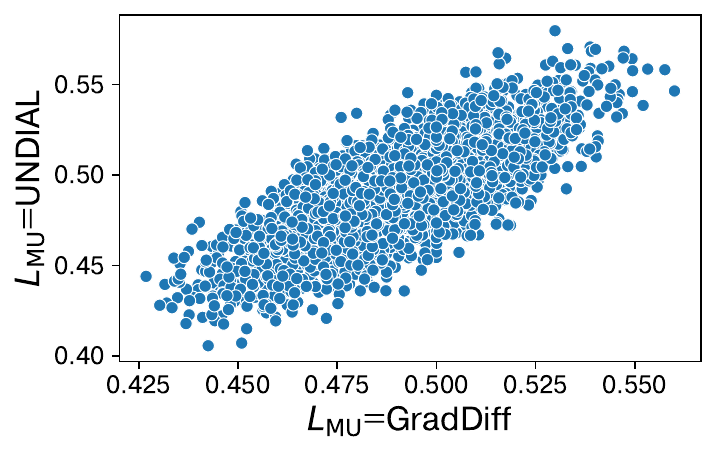}
    \caption{\metric score is robust to choices of $L_{\text{MU}}$ in Eq.~\ref{eq:find_easy} and \ref{eq:find_hard}, since the regularizer leads to sparse and consistent selection of representative samples. Correlation coefficient $\rho=0.76$. Each point represent in the figure represent the CUD scores computed using the GradDiff~\citep{maini2024tofu} MU loss and the UNDIAL~\citep{dong2024undial} MU loss.}
    \label{fig:cud_diff_loss}
\end{figure}

\paragraph{\metric is robust to choice of loss function:}
In identifying representative easy and hard circuits (Eq.~\ref{eq:find_easy} and \ref{eq:find_hard}), we adopt a simple formulation of the unlearning loss $L_{\text{MU}}$ (GradDiff), while more sophisticated alternatives exist. We show that \metric remains stable across different loss choices. Specifically, we evaluate the loss function of UNDIAL.
Figure~\ref{fig:cud_diff_loss} presents the comparison of \metric scores computed under different choices of $L_{\text{MU}}$. The two scores show strong agreement, with a correlation coefficient $\rho=0.76$.
This robustness arises from the sparsity-inducing regularizer $\lambda \lVert w \rVert$, which constrains the selected circuit to remain compact. As a result, the resulting easy and hard sample sets are highly consistent across loss functions, leading to stable \metric values despite variations in the underlying objective.



\begin{table}[t]
\centering
\small
\setlength{\tabcolsep}{4pt}
\caption{CUD is not explained by superficial lexical information. The most salient n-grams are diverse and do not cluster around any coherent topic, suggesting that CUD does not simply reflect domain-specific lexical cues or fact-level semantic overlap.}
\label{tab:lexical}
\begin{tabular}{lr|lr}
\toprule
\textbf{Easy Unigram} & \textbf{Freq} & \textbf{Easy Bigram} & \textbf{Freq} \\
\midrule
What       & 33 & Hsiao Yun	         & 16 \\
writing    & 32 & writing style	     & 14 \\
work       & 30 & Tae ho	         & 14 \\
books      & 29 & Elvin Mammadov     & 12 \\
How        & 28 & Yun Hwa's          & 11 \\
genre      & 28 & Takashi Nakamura's & 11 \\
influenced & 26 & Xin Lee	         & 10 \\
works      & 26 & Ji Yeon	         &  9 \\
LGBTQ      & 23 & Nikolai Abilov's	 &  9 \\
author     & 22 & Wei Jun	         &  8 \\
\midrule
\midrule
\textbf{Hard Unigram} & \textbf{Freq} & \textbf{Hard Bigram} & \textbf{Freq} \\
\midrule
What      & 45 & Ji Yeon           & 20 \\
genre     & 45 & Kalkidan Abera    & 20 \\
books     & 35 & Hina Ameen        & 17 \\
author    & 31 & Yeon Park         & 16 \\
Al        & 26 & author born       & 13 \\
writing   & 21 & full name         & 11 \\
Kalkidan  & 21 & Hsiao Yun         & 11 \\
Abera     & 21 & Carmen Montenegro & 11 \\
born      & 20 & Jad Ambrose       & 10 \\
Ji        & 20 & Ambrose Al        & 10 \\
\bottomrule
\end{tabular}
\end{table}

\subsection{CUD Captures Mechanistic Unlearning Difficulty}

\paragraph{CUD is not confounded lexical similarity:}
A natural concern is that CUD may be driven by lexical artifacts associated with particular tokens, rather than by genuinely mechanistic differences. To examine this, we analyze the most salient unigram and bigrams contributing to the aggregated easy and hard circuits. As shown in Table~\ref{tab:lexical}, the top-ranked unigram and bigrams are diverse and do not correspond to any coherent topical category, indicating that the anchors do not preferentially encode domain-specific information. This is because anchor points in CUD are aggregated circuits of multiple samples, which are designed to capture shared mechanistic properties of internal computation reflect structural patterns in model processing, rather than semantic similarity to particular facts.

\paragraph{CUD is not confounded by context length:}
A another concern is that CUD may be confounded by context length. Longer facts may require more tokens to express and therefore produce circuits that look systematically different from those of short, single-token facts (for example, ``Paris'' as the capital of France), regardless of their true unlearning difficulty. We therefore test whether CUD is biased by context length.
We plot CUD scores against context length across all evaluated samples and compute their correlation. We observe no meaningful association ($\rho = -0.02$): both short and long inputs exhibit a wide spread of CUD scores. This result suggests that CUD does not simply track context length, but instead captures deeper mechanistic differences that are relevant to unlearning difficulty.

\begin{figure}
    \centering
    \includegraphics[width=0.98\linewidth]{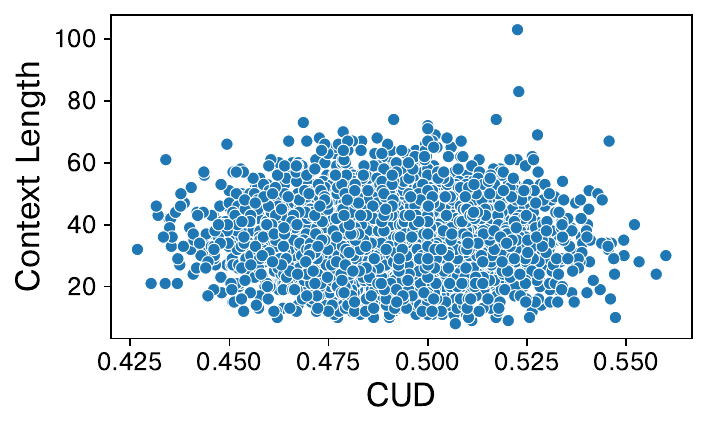}
    \caption{CUD is not confounded by context length. Across evaluated samples, CUD scores exhibit no meaningful correlation with input length ($\rho=-0.02$), and both short and long contexts span a wide range of values, suggesting that CUD does not merely track token count or superficial circuit size.}
    \label{fig:cud_vs_context_length}
\end{figure}

\subsection{Mapping Unlearning Difficulty to Internal Model Mechanism}


\paragraph{Distinct Edge Distribution:} The performance disparity between easy and hard samples is explained by differences in the structural differences between easy and hard circuits. 

Figure~\ref{fig:edge_dist} overlays the histogram of circuit-edge usage for the \textcolor{blue}{Easy} and \textcolor{red}{Hard} samples. Circuit edges are ordered by decreasing frequency on the $x$-axis, while the $y$-axis reports the number of times each edge appears across the extracted circuits. The distribution of easy and hard edges are statistically different, with a $p$-value of $0.01$. Importantly, the distribution of easy edges shows systematically higher counts in the head of the distribution, indicating stronger concentration on a small set of dominant edges. In contrast, the distribution of hard edges is comparatively flatter, with fewer highly reused edges and more mass distributed across low-frequency edges. This suggests that easy-to-unlearn behavior is associated with compact, repeatedly utilized computation paths, while hard-to-unlearn behavior draws on more heterogeneous and diffuse circuitry.
One common part is that both easy and hard edges exhibit a heavy-tailed profile -- a small subset of edges is reused many times, whereas the majority of edges occur rarely. 

\looseness=-1 The performance disparity between easy and hard samples is explained by differences in the \emph{underlying circuits} that implement their predictions. Easy samples are mediated by a small, high-frequency sub-circuit (shared edges with large reuse), implying their behavior depends on a relatively low-dimensional and stable set of mechanisms. Consequently, unlearning can succeed by disrupting a limited number of dominant edges, yielding large behavioral change with localized intervention. Hard samples, by contrast, rely on a broader set of low-frequency edges, consistent with \emph{redundant} and \emph{entangled} representations spread across multiple components. Forgetting such samples requires coordinated changes across many edges, making them intrinsically more resistant to unlearning. In this view, difficulty is not merely a data-side phenomenon but a mechanistic one: compact and reusable circuits tend to be easier to erase, whereas distributed circuits are harder to remove without collateral damage to retained behavior.

\begin{figure}
    \centering
    \includegraphics[width=0.98\linewidth]{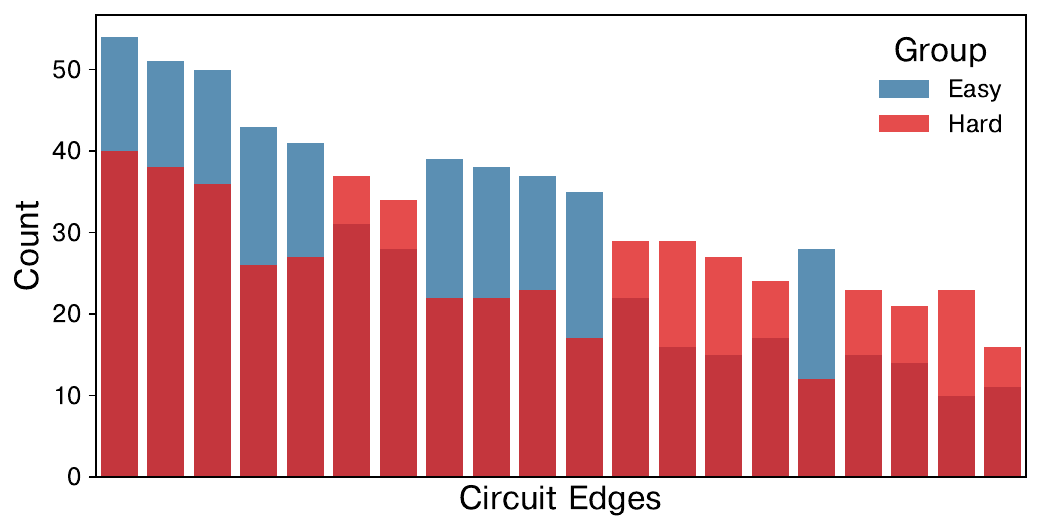}
    \caption{Edge distribution of statistically different easy and hard circuits.}
    \label{fig:edge_dist}
\end{figure}

\paragraph{Top Unique Edges:}
Table~\ref{tab:top_edges} in Appendix~\ref{sec:additional_results} reports the top-10 edges that appear more frequently in easy/hard circuit.
The edges that appear more frequently in easy samples are primarily concentrated in early-to-mid MLP pathways, including direct input injections (\eg $\text{input} \rightarrow \text{m0}$, $\text{input} \rightarrow \text{m1}$) and local MLP-to-MLP transitions such as $\text{m0} \rightarrow \text{m2}$, $\text{m1} \rightarrow \text{m4}$, and repeated fan-out from a single layer (e.g., $\text{m2} \rightarrow \text{m3}$, $\text{m2} \rightarrow \text{m5}$, $\text{m2} \rightarrow \text{m6}$, $\text{m2} \rightarrow \text{m8}$). These edges largely remain within the feed-forward stack and do not directly interact with the output layer, indicating that easy samples are resolved through relatively shallow, modular transformations that propagate smoothly from the input to intermediate representations.

In contrast, edges that occur more frequently in hard samples are skewed toward late-stage and output-facing circuits. This set is dominated by deeper MLP transitions (e.g., $\text{m6} \rightarrow \text{m11}$, $\text{m11} \rightarrow \text{m13}$, $\text{m11} \rightarrow \text{m15}$) and direct connections to the logits (e.g., $\text{m9} \rightarrow \text{logits}$, $\text{m10} \rightarrow \text{logits}$), suggesting stronger reliance on high-level features that are tightly coupled to the model’s final decision process. Moreover, the presence of attention-mediated routing, such as $\text{m6} \rightarrow \text{a7.h2}\langle v \rangle$, highlights an additional layer of representational integration that is absent from the easy circuits.

Overall, these patterns point to a clear mechanistic distinction: easy samples are uniquely supported by shallow MLP edges anchored close to the input and intermediate part, whereas hard samples depend on deeper, output-proximal and attention-involving pathways. This structural shift toward late-layer aggregation provides a plausible interpretation for why hard samples exhibit greater resistance to unlearning.\looseness-1

\subsection{Scalability of CUD}

We study the scalability of time spent on computing anchors. Results in Appendix~\ref{sec:additional_results} Table~\ref{tab:scale} show that anchor construction runtime scales linearly with respect to the forget size.
Similarly, computation time scales linearly with model size.
Therefore, anchor construction runtime is substantially better than quadratic growth, which supports practical use at larger scales.

\section{Discussion}

We position \metric relative to several existing perspectives on unlearning difficulty at sample and group levels. The key novelty of \metric is that it provides the first \emph{continuous, pre-unlearning, circuit-grounded} notion of per-sample difficulty, which enables both quantitative stratification and mechanistic explanation.

\subsection{Adversarially Challenging Forget Set}
Adversarially challenging (worst-case) forget set selects a small set of adversarial samples whose loss remains low after unlearning, \ie failed to unlearn, through optimization~\citep{fan2024challenging}. 

\looseness=-1\xhdr{\metric is finer-grained} Worst-case selection~\citep{fan2024challenging}
is binary, focusing exclusively on the most extreme samples: each sample is either included in the forget set or not. In contrast, \metric assigns a continuous difficulty score, preserving fine-grained information. This enables richer analyses, such stratifying results by difficulty range and studying correlation between difficulty and outcomes such (e.g., unlearning efficacy).

\looseness=-1\xhdr{Mechanistic debugging} Worst-case forget set provides little insight into why samples resist forgetting. In contrast, circuit-based \metric grounds unlearning difficulty in the internal mechanisms used by the model for decision making. This enables fine-grained analysis of which layers, modules, or circuit communities dominate resistance to forgetting, facilitating targeted debugging and intervention. For example, resistance localized to late-layer MLP circuits suggests different unlearning strategies than resistance mediated by early attention pathways.

\xhdr{Potential future applications} The continuous nature of \metric enables a range of difficulty-aware unlearning strategies that are not supported by binary worst-case sets. This opens the door to smooth curricula or pacing strategies (e.g., easy-to-hard), difficulty-aware sampling, loss reweighting, or constrained selection strategies. 
We leave the exploration of these applications to future work.




\subsection{Memory Removal Difficulty}

Memory Removal Difficulty (MRD)~\citep{feng2025neuro} is a neuroscience-inspired measure of sample-level unlearning difficulty, which quantifies how sensitive a sample’s likelihood is to small perturbations of model parameters. Samples whose likelihood remains largely unchanged under perturbations are considered heavily memorized and therefore hard to unlearn, whereas samples with larger likelihood shifts are deemed easier to unlearn. The default range of MRD is in [0,2], with 1 - MRD/2 as a normalized difficulty score in [0, 1].

Figure~\ref{fig:cud_vs_mrd} shows the relationship between \metric and MRD-based difficulty (\ie 1 - MRD/2) for all samples in TOFU. Overall, the two metrics exhibit a weak correlation ($\rho=-0.27$), indicating that they capture fundamentally different notions of unlearning difficulty. While \metric produces a roughly normal and balanced distribution, MRD assigns a large fraction of samples very small scores, effectively categorizing most samples as hard to forget.
For a narrow range of \metric values, MRD spans a wide range. This dispersion indicates that MRD is highly sensitive to fine-grained parameter fluctuations, whereas \metric captures mechanistic structure that directly influences the model’s decision-making behavior. In other words, MRD reflects local instability under perturbations, while \metric encodes functionally relevant mechanisms that govern how predictions are formed. 

Moreover, MRD can only be meaningful if computed online as unlearning proceeds, since it relies on perturbations on the immediate version of the trained model. While \metric can probe unlearning difficulty prior to any unlearning intervention, providing a predictive and method-agnostic characterization, a unique advantage of \metric.


\begin{figure}
    \centering
    \includegraphics[width=0.98\linewidth]{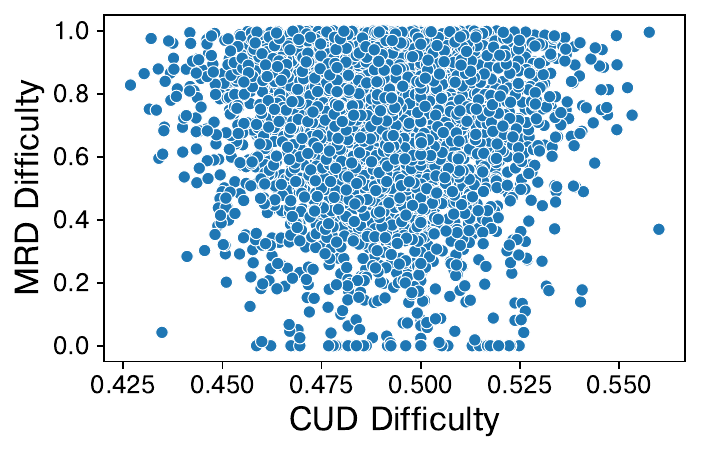}
    \caption{Comparison between \metric-based and MRD-based difficulty score on TOFU. \metric and MRD captures fundamentally different information when quantifying sample unlearning difficulty, with a correlation coefficient $\rho=-0.27$.}
    \label{fig:cud_vs_mrd}
\end{figure}

\subsection{Other post-hoc Analysis}

Recent post-hoc studies on image classification tasks highlight that not all samples are equally easy to forget~\citep{asami-sugawara-2024-makes,rizwan2024instance}. \citet{zhao2024what} argues the forget-retain entanglement and extent of memorization influence unlearning difficulty. Through instance-level analysis, \citet{rizwan2024instance} discover four empirical factors to explain why certain samples remain persistent: 1) proximity to the decision boundary, 2) resistance to membership inference, 3) number of unlearning steps, and 4) size of unlearning expansion.\looseness-1

While this line of work provides valuable post-hoc insights, these difficulty indicators are inherently observed {\em after} unlearning has been performed, relying on unlearning dynamics or attack outcomes to characterize difficulty. In contrast, our approach directly scores unlearning difficulty {\em prior} to unlearning, using mechanistic signals extracted from the model itself, enabling proactive identification of easy- and hard-to-unlearn samples without executing unlearning procedures. This distinction is critical in practice, as it allows difficulty-aware unlearning strategies, adaptive resource allocation, and principled benchmarking without incurring the cost of repeated unlearning trials. 


\section{Conclusion}

We investigate a key yet underexplored question in machine unlearning: {\em why unlearning difficulty varies substantially across samples}. 
We introduce {\em Circuit-guided Unlearning Difficulty} (CUD) score, a circuit-based metric that quantifies sample-level unlearning difficulty {\em prior} to any unlearning intervention. Across extensive experiments, CUD reliably separates easy and hard samples, and remains stable across unlearning methods. 
In addition, our circuit analyses suggest that unlearning difficulty is fundamentally tied to the internal mechanisms that support a model’s decision-making process.
%
%
We hope this work catalyzes research on mechanistic, circuit-level foundations of unlearning, and promotes the development of difficulty-aware unlearning methods that adapt to intrinsic sample difficulty rather than treating all samples uniformly.

Promising future directions include using CUD to 
(i) construct controlled forget sets at specified difficulty levels for benchmarking, 
(ii) design curriculum-style unlearning schedules (e.g., easy-to-hard or hard-focused pacing), 
(iii) develop adaptive sampling and loss reweighting models based on predicted difficulty, 
(iv) guide targeted interventions by localizing unlearning to specific layers or circuits, and 
(v) developing more efficient approximations of CUD, as well as lightweight proxies that preserve its predictive and mechanistic fidelity.

\section*{Limitations}
A limitation of our work is that CUD may be computationally expensive, since CUD requires circuit discovery. As a result, CUD is not intended to be used as an online or per-iteration diagnostic for the entire retain set, though acceptable for the forget set. It is best suited for offline analysis and pre-unlearning assessment, where interpretability is prioritized. In practice, this cost can be amortized by reusing discovered circuits across samples or caching intermediate representations. We view efficiency improvements and approximations of CUD as an important direction for future work.

\section*{Ethical Considerations}
This work aims to improve the interpretability and transparency of machine unlearning. By providing a principled way to analyze why certain samples are difficult to unlearn, our approach supports more accountable and explainable unlearning systems, which is essential for real-world deployment under legal and ethical constraints. All experiments are conducted on publicly available datasets, and no personally identifiable, sensitive, or private information is used. While mechanistic analysis can reveal internal model behaviors, we believe this increased transparency aligns with responsible AI practices and does not introduce new misuse risks beyond those already present in standard interpretability research.

\nocite{cheng2023multimodal}
\nocite{chen2025future}
\nocite{cheng2024mubench}
\nocite{cheng25d_interspeech}
\bibliography{reference,anthology}

\newpage
\appendix
\section{Original Models.}\label{sec:ori_model}
We use the following original models for each unlearning method in Table~\ref{tab:ori_model}.

\textbf{\texttt{GradDiff}} is a gradient-difference–based method that enforces forgetting by explicitly driving parameter updates in directions that reduce the influence of forget samples relative to retain data. It operates directly in parameter space and is sensitive to how gradients from forget examples are represented internally.

\textbf{\texttt{NPO}} follows the Negative Preference Optimization framework, which discourages the model from assigning high likelihood to forget samples while preserving performance on retained data. NPO implicitly reshapes decision boundaries through preference reweighting.

\textbf{\texttt{SimNPO}} extends NPO by incorporating similarity-aware constraints, encouraging the model to suppress forget samples while maintaining consistency for semantically similar retain examples. This introduces an additional structural bias into the unlearning dynamics.

\textbf{\texttt{RMU}} performs Representation-level Model Unlearning by selectively perturbing internal activations at designated layers. Rather than operating purely on outputs or losses, RMU intervenes at intermediate representations, making it particularly relevant for circuit-level analysis.

\textbf{\texttt{UNDIAL}} is a dialogue-aware unlearning method that balances forgetting and retention via constrained optimization. It explicitly trades off forget suppression and retain preservation through dual objectives, resulting in distinct internal adaptation patterns.

The original models are taken from a comprehensive LLM unlearning benchmark open-unlearning~\citep{dorna2025openunlearning}.

For LLMRec unlearning, we take the original models from ~\citet{wang2025towards}.

\section{Additional Results}\label{sec:additional_results}
Table~\ref{tab:scale} demonstrates that CUD can scale to larger forget sets and model sizes without introducing significant computation overhead.

\begin{table}[t]
\centering
\small
\setlength{\tabcolsep}{4pt}
\caption{Scalability of computing anchors.}
\label{tab:scale}
\begin{tabular}{rr}
\toprule
\textbf{Forget size (\%)} & \textbf{Time (min)} \\
\midrule
 2 &  14.3 \\
 4 &  23.7 \\
 6 &  35.2 \\
 8 &  46.6 \\
10 &  60.3 \\
20 & 103.5 \\
\midrule
\midrule
\textbf{Model Size (B)} & \textbf{Time (min)} \\
\midrule
1 &  60.3 \\
3 &  80.8 \\
8 & 125.3 \\
\bottomrule
\end{tabular}
\end{table}

\begin{table*}[t]
\centering
\small
\caption{Unlearning models evaluated in this work and their corresponding original model checkpoints.}
\label{tab:ori_model}
\begin{tabular}{l l}
\toprule
\textbf{Model} & \textbf{Original model} \\
\midrule
\texttt{GradDiff} & \texttt{open-unlearning/unlearn\_tofu\_Llama-3.2-1B-Instruct\_forget10\_GradDiff\_lr1e-05\_alpha5\_epoch10} \\

\texttt{NPO} & \texttt{open-unlearning/unlearn\_tofu\_Llama-3.2-1B-Instruct\_forget10\_NPO\_lr1e-05\_beta0.5\_alpha1\_epoch10} \\

\texttt{SimNPO} & \texttt{open-unlearning/unlearn\_tofu\_Llama-3.2-1B-Instruct\_forget10\_SimNPO\_lr5e-05\_b3.5\_a1\_d1\_g0.25\_ep5} \\

\texttt{RMU} & \texttt{open-unlearning/unlearn\_tofu\_Llama-3.2-1B-Instruct\_forget10\_RMU\_lr2e-05\_layer10\_scoeff100\_epoch5} \\

\texttt{UNDIAL} & \texttt{open-unlearning/unlearn\_tofu\_Llama-3.2-1B-Instruct\_forget10\_UNDIAL\_lr0.0001\_beta10\_alpha2\_epoch10} \\
\bottomrule
\end{tabular}
\end{table*}

\begin{table*}[t]
\small
\centering
\caption{CUD is robust to the choice of similarity metric. Under the same unlearning settings, hard set has lower Unlearning efficacy, retain performance, and general knowledge, indicating greater resistance to forgetting, whereas the easy set achieves higher performance across all metrics. Default set: the default forget/retain split on TOFU. Hard set: Hard forget set selected by CUD. Similar for Easy set. Numbers in parenthesis report the gap to default set and $p$-value of difference, respectively.}
\label{tab:tofu_new_split_comparison}
\begin{tabular}{p{4em}|l|ccc}
\toprule
\textbf{Sim Metric} & \textbf{Choice of $\mathcal{D}_f$} & \textbf{Unlearn Efficacy ($\uparrow$)} & \textbf{Retain ($\uparrow$)} & \textbf{General Knowledge ($\uparrow$)} \\
\midrule
Prior-Unlearn & - & 22.0 & 79.3 & 81.2 \\
\midrule
 & Default Set & 57.8 & 66.7 & 75.5 \\
\cmidrule{2-5}
\multirow{3}{*}{\textbf{Cosine}} 
& Hard Set by \metric & 43.7 (\textcolor{lightred}{-14.1}) (1e-15) & 64.3 (\textcolor{lightred}{-2.4}) (1e-4) & 73.1 (\textcolor{lightred}{-2.4}) (1e-4) \\
& Easy Set by \metric & 61.1 (\textcolor{lightgreen}{+3.3}) (1e-3) & 68.0 (\textcolor{lightgreen}{+1.3}) (1e-3) & 75.3 (\textcolor{lightred}{-0.2}) (1e-1) \\
\midrule
\multirow{3}{*}{\textbf{Jaccard}} 
& Hard Set by \metric & 45.0 (\textcolor{lightred}{-12.8}) (1e-13) & 63.9 (\textcolor{lightred}{-2.8}) (1e-4) & 73.9 (\textcolor{lightred}{-1.6}) (1e-3) \\
& Easy Set by \metric & 61.5 (\textcolor{lightgreen}{+3.7}) (1e-4) & 68.5 (\textcolor{lightgreen}{+1.8}) (1e-3) & 75.7 (\textcolor{lightgreen}{+0.2}) (1e-1) \\
\midrule
\multirow{3}{*}{\textbf{Hamming}} 
& Hard Set by \metric & 47.2 (\textcolor{lightred}{-10.6}) (1e-13) & 62.6 (\textcolor{lightred}{-4.1}) (1e-5) & 74.0 (\textcolor{lightred}{-3.9}) (1e-4) \\
& Easy Set by \metric & 60.2 (\textcolor{lightgreen}{+2.4}) (1e-3) & 67.5 (\textcolor{lightgreen}{+0.8}) (1e-2) & 75.2 (\textcolor{lightred}{-0.3}) (1e-1) \\
\bottomrule
\end{tabular}
\end{table*}

Table~\ref{tab:tofu_new_split_comparison} shows that \metric is robust to the choice of similarity metric used in its construction. 
Table~\ref{tab:rec_new_split_gpt2}--\ref{tab:rec_new_split_llama} demonstrates that CUD can select easy and hard forget sets on LLM Rec unlearning. 
Table~\ref{tab:top_edges} reports the top-10 edges that appear more frequently in easy and hard circuits.

\begin{table*}[t]
\small
\centering
\caption{Using CUD to select easy and hard set on LLM Rec unlearning with GPT2. Numbers in parenthesis report the gap to default set and $p$-value of difference, respectively.}
\label{tab:rec_new_split_gpt2}
\begin{tabular}{p{4em}|l|ccc}
\toprule
\textbf{Unlearn Method} & \textbf{Choice of $\mathcal{D}_f$} & \textbf{Unlearn Efficacy ($\uparrow$)} & \textbf{Retain ($\uparrow$)} & \textbf{General Knowledge ($\uparrow$)} \\
\midrule
\multirow{3}{*}{Average} 
& Default Set & 76.6 & 69.2 & 1.91 \\
\cmidrule{2-5}
& Hard Set by \metric & 56.2 (\textcolor{lightred}{-20.4}) *** & 53.3 (\textcolor{lightred}{-15.9}) *** & 2.21 (\textcolor{lightgreen}{-0.3}) ** \\
& Easy Set by \metric & 83.5 (\textcolor{lightgreen}{+6.9}) *** & 80.9 (\textcolor{lightgreen}{+11.7}) *** & 1.67 (\textcolor{lightred}{-0.24}) ** \\
\bottomrule
\end{tabular}
\end{table*}

\begin{table*}[t]
\small
\centering
\caption{Using CUD to select easy and hard set on LLM Rec unlearning with Llama3. Numbers in parenthesis report the gap to default set and $p$-value of difference, respectively.}
\label{tab:rec_new_split_llama}
\begin{tabular}{p{4em}|l|ccc}
\toprule
\textbf{Unlearn Method} & \textbf{Choice of $\mathcal{D}_f$} & \textbf{Unlearn Efficacy ($\uparrow$)} & \textbf{Retain ($\uparrow$)} & \textbf{General Knowledge ($\uparrow$)} \\
\midrule
\multirow{3}{*}{Average} 
& Default Set & 78.2 & 69.2 & 1.91 \\
\cmidrule{2-5}
& Hard Set by \metric & 58.3 (\textcolor{lightred}{-19.9}) *** & 53.2 (\textcolor{lightred}{-15.9}) *** & 2.21 (\textcolor{lightgreen}{0.2}) * \\
& Easy Set by \metric & 83.7 (\textcolor{lightgreen}{+5.5}) *** & 81.8 (\textcolor{lightgreen}{+12.6}) *** & 1.61 (\textcolor{lightred}{-0.3}) (1e-1) \\
\bottomrule
\end{tabular}
\end{table*}

\begin{table}[t]
\centering
\small
\setlength{\tabcolsep}{4pt}
\caption{Top edges unique to easy and hard circuit.}
\label{tab:top_edges}
\begin{tabular}{p{3.5em}p{6em}|p{4em}p{4em}p{2.2em}}
\toprule
\textbf{Edge ID} & \textbf{Edge} & Freq Easy (\%) & Freq Hard (\%) & $\Delta$ \\
\midrule
\multicolumn{5}{l}{Unique edges in easy circuit}\\
\midrule
135315 & m2$\rightarrow$m6     & 46.6 & 22.6 & 24.0 \\
135218 & m2$\rightarrow$m5     & 57.3 & 34.6 & 22.6 \\
48370  & m0$\rightarrow$m2     & 52.0 & 29.3 & 22.6 \\
135509 & m2$\rightarrow$m8     & 37.3 & 16.0 & 21.3 \\
193    & input$\rightarrow$m1  & 50.6 & 29.3 & 21.3 \\
241240 & m5$\rightarrow$m7     & 66.6 & 48.0 & 18.6 \\
135024 & m2$\rightarrow$m3     & 54.6 & 36.0 & 18.6 \\
93443  & m1$\rightarrow$m4     & 49.3 & 30.6 & 18.6 \\
96     & input$\rightarrow$m0  & 72.0 & 53.3 & 18.6 \\
209068 & m4$\rightarrow$m6     & 68.0 & 50.6 & 17.3 \\
\midrule
\multicolumn{5}{l}{Unique edges in hard circuit}\\
\midrule
270599 & m6$\rightarrow$m12           & 13.3 & 30.6 & -17.3 \\
174180 & m3$\rightarrow$m10           & 21.3 & 38.6 & -17.3 \\
338307 & m9$\rightarrow$logits        & 20.0 & 36.0 & -16.0 \\
270502 & m6$\rightarrow$m11           & 20.0 & 30.6 & -10.6 \\
367245 & m11$\rightarrow$m15          & 18.6 & 28.0 & -9.3 \\
383380 & m13$\rightarrow$m15          & 29.3 & 38.6 & -9.3 \\
367051 & m11$\rightarrow$m13          & 22.6 & 32.0 & -9.3 \\
354377 & m10$\rightarrow$logits       & 37.3 & 45.3 & -8.0 \\
318550 & m8$\rightarrow$m10           & 41.3 & 49.3 & -8.0 \\
270084 & m6$\rightarrow$a7.h2$\langle$v$\rangle$ & 14.6 & 21.3 & -6.6 \\
\bottomrule
\end{tabular}
\end{table}

\end{document}